\documentclass[discussion]{clv2025}
\usepackage{xcolor}
\usepackage{hyperref}

\jvol{vv}
\jnum{nn}
\jyear{2025}
\dochead{Last Words} 
\pageonefooter{Action editor: Saif Mohammad. Submission received: 27 April 2025; revised version received: 19 June 2025; accepted for publication: : 22 June 2025.}
\usepackage{amsmath}
\usepackage{booktabs}

\newcommand{\lingform}[1]{\textit{\texttt{#1}}}

\newcommand{\scare}[1]{`#1'}

\newcommand{\revised}[1]{{#1}}

\runningtitle{We Should Evaluate Real-World Impact}
\runningauthor{Reiter}

\begin{document}

\title{We Should Evaluate Real-World Impact}

%\author{Ehud Reiter}
\author{Ehud Reiter\thanks{}}

\affilblock{
    \affil{University of Aberdeen, UK \\\quad \email{e.reiter@abdn.ac.uk}}
}

\maketitle

\begin{abstract}
The ACL community has very little interest in evaluating the real-world impact of NLP systems.  A structured survey of the ACL Anthology shows that perhaps 0.1\% of its papers contain such evaluations; furthermore most papers which include impact evaluations present them very sketchily and instead focus on metric evaluations.  NLP technology would be more useful and more quickly adopted if we seriously tried to understand and evaluate its real-world impact.
\end{abstract}

\section{Introduction}

The medical community places great emphasis on clinical trials that assess the real-world effectiveness of new medications and other interventions; papers in education and engineering also regularly present data on real-world impact of new techniques.  But in NLP, very few papers (even in Industry or Applications tracks) present data on real-world effectiveness of deployed systems.  We regularly claim that LLMs and other NLP technologies are changing the world \cite{aiindex2025}, but we are reluctant to provide data on how deployed NLP systems improve real-world  \textit{key performance indicators} (KPIs); \revised{I will refer to such changes in KPIs as \textit{impact} below}. 

Of course providing data on changes in real-world KPIs is hard, much harder than calculating benchmark and metric scores on a test set, or asking crowdworkers to assess the quality of an output.  Measuring impact requires deploying the system in production usage, monitoring its effect on users, and getting permission to publicly release this data.  But still, other fields are publishing papers on real-world impact of LLMs, including medicine \citep{Duggan2025} and software engineering \citep{pandey2024}, so it can be done.

It is also notable that even when real-world impact is measured, many NLP papers treat this as secondary to a metric evaluation.  For example, \citet{maheshwary-etal-2024-pretraining} carefully describe a metric-based evaluation, and then in a single paragraph summarise a real-world A/B impact evaluation which shows significant improvements in important KPIs (Section~\ref{sec:industry}).  Similarly \citet{yoon-etal-2024-language} carefully describe a metric-based evaluation, and then briefly mention that a real-world before-and-after study showed very impressive KPI improvements.  In short, many NLP researchers do not seem to think real-world impact is important (at least in an academic paper), even when they have data on this.
%even when data on real-world impact is available, many NLP authors do not prioritise communicating this information in their papers.
%What is especially frustrating, at least to me, is papers that do report on interesting real-world impact data, but in a very sketchy way which makes it difficult to understand what the experiment really showed.  For example, \citet{maheshwary-etal-2024-pretraining} use 2 pages in their paper to carefully describe a metric-based evaluation, and then in a single paragraph mention a real-world A/B impact evaluation which shows significant improvements in important KPIs.  Similarly \citet{yoon-etal-2024-language} carefully describe a metric-based evaluation, and then briefly mention that a real-world before-and-after study showed very impressive KPI improvements.  In short, even when data on real-world impact is available, many NLP authors do not prioritise communicating this information in their papers.

In this paper I first discuss and give examples of impact evaluation.  I then show in a literature survey that such papers are  rare in NLP, and conclude with a discussion and suggestions for encouraging more NLP researchers to evaluate real-world impact.

%Types of evaluation
%Define impact evaluation
%Why – assess genuine utility, group other eval methods

\section{Impact evaluation}\label{testset}

There are many ways of evaluating NLP systems.  Most evaluations involve running several systems (including a baseline) on a \emph{test set} of data or scenarios, and evaluating how well the NLP systems do on the test set.
Such evaluations are most commonly done using automatic metrics such as precision and recall for classification, BLEU score \citep{papineni-etal-2002-bleu} for text generation, and (more recently) using LLMs as judges \cite{zheng-etal23}.  Some studies use human evaluation; the most common form of this is asking human subjects to rate outputs or annotate problems in outputs \cite{reiter2024}.  Human evaluation can also be \emph{extrinsic} \cite{jones1995}, which involves measuring the effect the NLP system has on helping users do something, such as make good clinical decisions \citep{PORTET2009789}.

\revised{Such evaluations have been very successful in driving the development of core NLP technologies in areas such as speech recognition and machine translation \cite{liberman2020human}.  But they do not give complete understanding of how well NLP systems work in complex real-world applications.  Also, researchers may optimise test set performance in ways which do not actually increase real-world utility; this is an aspect of Goodhart's Law\footnote{\url{https://en.wikipedia.org/wiki/Goodhart\%27s\_law}} (\scare{When a measure becomes a target, it ceases to be a good measure}).}

An impact evaluation, in contrast, involves deploying a system and measuring changes in KPIs in real-world usage; since we are directly measuring the KPIs which we care about, Goodhart's Law is not a problem.  Note that while an impact evaluation can assess multiple systems,
%(for example, clinical trials or A/B tests)
different systems usually interact with different users.  So there is usually \emph{not} a single test set which is presented to all the systems being evaluated.

%Evaluating real-world impact is a type of \emph{extrinsic} evaluation \cite{jones1995}, that is an evaluation which measures the impact of an NLP system on a task.  Most extrinsic evaluations are done in artificial laboratory settings, sometimes as controlled psychological experiments (for example \citet{PORTET2009789}).  An impact evaluation is an extrinsic evaluation which is done in a real-world setting,
%(and hence has high \emph{ecological validity}),
%usually on deployed systems, and which measures KPIs (\emph{key performance indicators}) which are important to users or other stakeholders.

Two papers by Moramarco illustrate the difference between extrinsic evaluation on a test set and evaluating impact.  \citet{moramarco-etal-2022-human} present an extrinsic evaluation of systems which summarised clinician-patient consultations.  In this evaluation, they used a test set of mock consultations (consultations between a clinician and an actor) \cite{papadopoulos-korfiatis-etal-2022-primock57}. They asked clinicians to listen to the mock consultations, read computer-generated summaries from several different models (and also a human-written corpus text), and post-edit the summaries to make them accurate and acceptable; they measured the time needed to post-edit.  \citet{moramarcophd} did an impact evaluation of a single system in this domain. In this study Moramarco measured post-edit time when the final system was deployed and used with real patients in actual consultations (Section~\ref{moramarcophd}); doctors post-edited consultation summaries during or immediately after the consultation, as part of their normal clinical workflow.

\section{Examples of impact evaluation techniques}

Impact can be evaluated by comparing KPIs between people who use an NLP system and people who are in control group(s); techniques
for doing this include clinical trials, A/B testing, and before-and-after studies. Impact can also be evaluated in observational studies where KPIs are measured just in users of the NLP system.
I give examples below.

Note that impact can also be evaluated \emph{qualitatively}, using user observations and feedback, expert analyses of case studies, error analysis, example outputs, etc.  While the NLP community strongly emphasizes quantitative results, \citet{Kapoor_Henderson_Narayanan_2024} argue that in an AI in Law context, good qualitative studies are more meaningful than quantitative studies.  Qualitative observations can be also be very important in building good theoretical models. Regardless, since qualitative analyses of impact in NLP are usually supplements to quantitative analyses, I will focus on quantitative analyses in this paper.

\subsection{Clinical trials}

\revised{Clinical trials are popular in medicine.  In such experiments, patients are allocated to different groups, which receive different treatments; KPIs are measured and compared between groups.  There are many types of clinical trials \citep{greenhalgh1997read};
%\footnote{\url{https://en.wikipedia.org/wiki/Clinical\_study\_design}}
the most highly regarded are randomised (patients are randomly allocated to groups) and blind (patients do not know which group they are in).}

The STOP project \citep{reiter-etal-2001} built and evaluated a Natural Language Generation system which generated letters giving advice on smoking cessation, using data from a smoking questionnaire which participants filled out.  The evaluation was a randomised controlled clinical trial:
\begin{itemize}
    \item We recruited 2553 smokers, and randomly allocated them to three groups of roughly equal size: STOP, FixedLetter, ThankYou.
    \item All smokers filled out the smoking questionnaire.
    \item Smokers in STOP group were sent STOP letters, smokers in FixedLetter group were sent a fixed letter (manually edited version of STOP's default letter), and smokers in ThankYou group were just a letter thanking them for being in our study.
    \item After six months, we asked smokers if they had stopped smoking (this was our KPI); we verified this with a saliva test for nicotine residues.
\end{itemize}
Note this was not a blind trial, since patients knew which group they were in.

Unfortunately, the results showed that STOP was not effective; cessation rates were in fact highest in the FixedLetter group, although the difference was not statistically significant.

For more information on the STOP experiment, see \citet{Lennox2001} and \citet{reiter-etal-2003}.

\subsection{A/B testing}

\revised{A/B testing is a common technique used for evaluating web page and resources.  Users coming to the web page are randomly directed to either a baseline or new page, and their behaviour is monitored.  In a sense, A/B testing adapts clinical trial methodology to non-clinical use cases, including sales and marketing.}

\citet{russell-gillespie-2016-measuring} used A/B testing to evaluate whether a customised machine translation (MT) system did better than a generic MT system when translating user reviews on an e-commerce website.  The evaluation was structured as follows:
\begin{itemize}
    \item 88,106 visitors to an e-commmerce website who needed translations were randomly allocated to GenericMT or CustomMT groups.
    \item Visitors in  GenericMT group saw the output of the generic MT system; visitors in the CustomMT group were shown the output of the custom MT system.
    \item Researchers measured three KPIs: pages per visit, whether items were added to a cart, and whether items were bought (conversion rate).
\end{itemize}
Results showed that the CustomMT group had (statistically significant) higher values for all KPIs; for example conversion rate increased by 8.7\%.

\subsection{Before-and-after studies}\label{moramarcophd}
\revised{A \emph{before-and-after} study (sometimes called \emph{pre-post study}) measures changes in KPIs after a new tool is introduced.  This means that KPIs based on the new tool are measured after pre-tool KPIs, which is not ideal (its better to measure both at the same time).  However, such analyses have the advantage that they may not require a special experiment to be organised, they can instead be based on data collected during routine operations.}

Unfortunately, I am not aware of any paper in the ACL Anthology which reports before-and-after studies at an acceptable level of detail.  A before-and-after study is described in detail in Chapter 7 of Moramarco's PhD thesis \cite{moramarcophd}.  This evaluated the impact of a tool that generated summaries of doctor-patient consultations (which are needed for the patient record).  The evaluation was structured as follows:
\begin{enumerate}
    \item 20 clinicians were selected, and the time they spent writing summaries in real consultations (before the tool was deployed) was calculated.
    \item The summarisation system was deployed, and the time these clinicians spent editing the computer-generated summaries was calculated, again in real consultations.  This was done over a 7-month period.  Comparing this to (1) measures changes in the KPI of time spent creating the summary.
    \item 20 manually-written summaries and 20 edited-computer-summaries (from the same clinicians) were carefully checked for errors.  This measured changes in the KPI of number of mistakes in summaries.
\end{enumerate}
Results were positive but not overwhelmingly so.  Post-editing computer summaries was 9\% faster than manually writing summaries.
\revised{This is relatively small, so the tool may not be cost-effective from a \textit{Return on Investment} (\textit{ROI}) perspective.}
%(ie, from a commercial perspective, the benefits from a small decrease in reporting time may not justify the expense of developing, deploying, and maintaining the NLP system).
There were also slightly fewer mistakes in the posted-edited summaries.

\subsection{Observational studies}
\revised{The techniques described above all compare an NLP system to something else, and report changes in KPIs. But if an NLP system is doing a novel task which has not been done before, then comparison is difficult and the paper may just report observed KPIs without comparisons.}

For example, \citet{nygaard-etal-2024-news} describe an NLP system for a credit officers which alerts them to relevant news about companies which have borrowed money.  The system was evaluated as follows:
\begin{itemize}
    \item It was deployed for two types of alerts, sentiment and mergers/acquisitions.
    \item 3500 alerts produced by the system were manually classified as new information, recently considered (already known), irrelevant, etc.
\end{itemize}
23\% of the alerts were new information, and hence useful to the credit officers.

\section{Literature Search}

In order to get a better understanding of impact evaluations in NLP venues, I performed two structured literature searches of the ACL Anthology.  The first was on papers in the EMNLP24 Industry track; this gave me a better understanding of keywords which identified papers which reported real-world impact.  I then used these keywords to search the entire ACL Anthology.

I focused on papers in the ACL Anthology because my goal is to understand the NLP community's perspective on impact evaluation.
The medical literature contains papers which describe clinical trials of NLP systems (eg, \citet{MEYSTRE2008602}) and before-and-after studies of NLP systems (eg, \citet{Duggan2025}), but my focus is the NLP literature, not the medical one.

\subsection{Papers in EMNLP24 industry track}\label{sec:industry}

Industry tracks at xACL conferences solicit papers about real-world applications, so seem especially likely to include impact studies.  The largest industry track at the time of writing was at EMNLP 2024, which included 122 papers.  I read the abstracts of all 122 papers; if the abstract suggested any kind of impact evaluation, I read the paper body.  I excluded papers which described technical details of deployment but did not give impact data, such as \citet{singhal-etal-2024-geoindia}.

\begin{table}
    \begin{tabular}{|l|r|l|}
    \hline
    \textit{Type} & \textit{count} & \textit{detailed}  \\ \hline
    clinical trial & 0 & -- \\ \hline
    A/B test & 6 & 2 (33\%) \\ \hline
    before-after & 1 & 0 (0\%) \\ \hline
    observational & 3 & 1 (33\%) \\ \hline
    TOTAL & 10 & 3 (30\%) \\ \hline
    \end{tabular}
    \caption{Impact papers in EMNLP24 industry track.  \textit{detailed} means description of impact study in the paper was at least 0.5 pages and included at least one figure or table.}\label{tab-industry}
\end{table}

This resulted in 10 papers (8\% of total) which gave impact data; most of these used A/B testing.  Most of these papers gave very short descriptions of the impact study.
For example, \citet{maheshwary-etal-2024-pretraining} describe a metric study in detail, and then describe an A/B impact study in a few sentences as follows:
\begin{quote}
%The LOFI pipeline successfully automated various insurance claim documents process. Clients verified that our pipeline achieved an average accuracy of 97\% across different document types. This resulted in a reduction of processing time by over 60\% and a decrease in staff requirements by 40\%.    
%\end{quote}
\lingform{After observing significant improvements during
offline simulations, we launched an online A/B experiment on live traffic to determine the impact of our proposed approach on geocode learning. We performed the model dial-up in a phased manner — 10\%, 50\%, and 100\% traffic. We observed statistically significant improvements during one week of dial-up in each phase. During the A/B test period, our approach learnt geocodes for a few hundred thousand shipments, where we observed 14.68\% improvement in delivery precision and 8.79\% reduction in delivery defects.}
\end{quote}
Since many details are missing, it is difficult to interpret the impressive-sounding improvements in KPIs.

Because of this problem, I checked which papers (A) devoted at least half a page (one column) to the impact study and (B) included at least one table or figure from the study.  Only three of the ten papers met this criteria.  See Table~\ref{tab-industry} for more information.

%Other papers gave more detail about their impact evaluation, including  \citet{mohankumar-etal-2024-improving} and  \citet{nygaard-etal-2024-news}.  However
I was also disappointed that \emph{none} of the papers gave the kind of detailed information about the impact evaluation which is present in the above-mentioned medical papers \cite{MEYSTRE2008602, Duggan2025}).  These papers are shorter than full-length xACL papers, so the problem is not size limitations.
%9 of the 10 papers gave far more detail about their metric evaluation than their impact evaluation (the exception was \citet{nygaard-etal-2024-news}).

%Some descriptions were so limited that it was unclear what really happened.

%, but it was unclear whether this was based on real-world data or not.
%\citet{maheshwary-etal-2024-pretraining} described a metric study in detail, but then described their \scare{Online A/B Experiment} in a single paragraph, which left me uncertain whether they did actually an A/B test.  Similarly \citet{yoon-etal-2024-language} described a metric study in great detail, and then mentioned a use case (automation of claim document processing) which seemed to be a before-after impact study, but it was unclear whether this was based on real-world data or not.

%With regard to KPIs measured, perhaps not surprisingly these were varied and usecase-specific, ranging from click-through rate to delivery defects.

Anyways, one goal of this exercise was to identify keywords (for title and abstract) which could be used to search the rest of the Anthology for papers which included an impact evaluation. Unfortunately, terminology varied widely in the the Industry Track papers.   The best keywords I found were \scare{A/B test} and \scare{deployed}, but these only matched six of the ten papers.

In short, this analysis suggests that (at least in Industry Track papers):
\begin{itemize}
    \item The most common type of impact evaluation in NLP is A/B testing, and the second most common is observational studies.
    \item When an impact evaluation is presented in an NLP paper, it is usually secondary to a metric evaluation, and often described very briefly.  This is the case even in Industry Track papers written by people from companies.
    \item The best title/abstract keywords for identifying papers with impact evaluations are \scare{A/B test} and \scare{deployed}, but many papers do not mention either of these keywords in their title or abstract.
\end{itemize}

Incidentally, EMNLP24 as a whole included around 3000 papers (main conference, findings, workshops, industry track).  I am only aware of one EMNLP paper not in the Industry Track which presented impact studies \citep{dai-etal-2024-contrastive}, so overall 0.37\% (11/3000) of EMNLP24 papers included impact evaluation.

\subsection{Papers in ACL Anthology}

I downloaded the ACL Anthology bib file on 12 March 2025; it contained 105,850 papers.  I used Zotero to search for papers whose title or abstract contained \scare{deployed}, \scare{a/b test}, \scare{clinical trial}, \scare{before-and-after evaluation}, or \scare{pre-post evaluation}.  This resulted in 537 papers.

I checked these 537 papers and identified 41 papers which
 gave quantitative impact data from a deployed system.  As mentioned above, I excluded papers which gave purely qualitative data, such as \citet{varges-etal-2012-semscribe}.

So all together, if I include the 4 EMNLP24 Industry Track papers which did not use any of the above keywords, I found 45 papers in the ACL Anthology which included impact studies; this is 0.04\% of the papers in the Anthology.

Of course there may be additional papers in the Anthology which report impact but do not use any of the above keywords.  The stats from Section~\ref{sec:industry} (40\% of the ten papers did not use any of the keywords), suggest that the actual percentage of Anthology papers which include impact studies may be closer to 0.1\%.
\revised{This is an estimate, but my expectation is that the percentage of Anthology papers which include impact evaluation is between 0.05\% and 0.2\%, with 0.1\% being my best estimate.}

\revised{From a venue perspective:
\begin{itemize}
    \item 32 (71\%) of the papers appeared in Industry Tracks.
    \item 7 (16\%) of the papers appeared in xACL conferences (including EMNLP, Coling, and Findings) outside of Industry Tracks.
    \item None of the papers appeared in journals.
    \item 6 (13\%) of the papers appeared in other Anthology venues (mostly workshops).
\end{itemize}}

%32 (70\%) of the 45 papers appeared in Industry Tracks.  I cannot tell exactly how many papers appeared in the NLP Applications track (track information is not given in the Anthology), but it is probably 5 at most.  This agrees with \citet{ganesh-etal-2023-mind} observation that Application track papers are often not grounded in real-world contexts.

Tables \ref{tab-type} and \ref{tab-keyword} break down the 45 papers by study type and keyword.  Overall, this is similar to the results in Section~\ref{sec:industry}:
\begin{itemize}
    \item Very few Anthology papers include impact evaluations.
    \item A/B tests are the most common type of impact evaluation, followed by observational studies.
    \item Only a third of papers that report impact studies describe them in even moderate detail.
\end{itemize}
Table~\ref{tab-papers} lists the 14 papers I found in the Anthology which report impact studies in at least moderate detail.

\begin{table}
    \begin{tabular}{|l|r|l|}
    \hline
    \textit{Type} & \textit{count} & \textit{detailed}  \\ \hline
    clinical trial & 1 & 1 (100\%) \\ \hline
    A/B test & 37 & 10 (27\%) \\ \hline
    before-after & 1 & 0 (0\%) \\ \hline
    observational & 6 & 3 (50\%) \\ \hline
    TOTAL & 45 & 14 (31\%) \\ \hline
    \end{tabular}
    \caption{Number of impact papers found in ACL Anthology, by study type.  \textit{detailed} means description of impact study in the paper was at least 0.5 pages and included at least one figure or table.}\label{tab-type}
\end{table}

\begin{table}
    \begin{tabular}{|l|r|l|}
    \hline
    \textit{Keyword} & \textit{count} & \textit{detailed}  \\ \hline
    clinical trial & 1 & 1 (100\%) \\ \hline
    A/B test & 22 & 7 (32\%) \\ \hline
    before-after & 0 & -- \\ \hline
    deployed & 24 & 6 (25\%) \\ \hline
    pre-post evaluation & 0 & --- \\ \hline
    (\textit{EMNLP industry Track}) & 10 &  3 (30\%)\\ \hline
    TOTAL & 45 & 14 (31\%) \\ \hline
    \end{tabular}
    \caption{Number of impact papers found in ACL Anthology, by abstract/title keyword.  Some papers match multiple keywords, so TOTAL is less than the sum of per-keyword counts.  \textit{detailed} means the description of the impact study in the paper is at least 0.5 pages and includes at least one figure or table.}\label{tab-keyword}
\end{table}

\begin{table}
    \begin{tabular}{|l|l|l|} \hline
    \textit{Paper} & \textit{type} & \textit{application area} \\ \hline
    \citet{reiter-etal-2001} & clinical trial & smoking cessation \\ \hline
    \citet{russell-gillespie-2016-measuring} & A/B & machine translation \\ \hline
    \citet{elsafty-etal-2018-document} & A/B & recommender system \\ \hline
    \citet{chapman-etal-2020-natural} & observational & identifying COVID cases \\ \hline
    \citet{srivastava-etal-2021-pretrain} & A/B & customer service \\ \hline
    \citet{liao-fares-2021-practical} & observational & customer service \\ \hline
    \citet{joshi-etal-2022-augmenting} & A/B & multi-label classification \\ \hline
    \citet{golobokov-etal-2022-deepgen} & A/B & generating advertisements \\ \hline
%    \citet{lei-etal-2022-plato} & A/B & generating advertisements \\ \hline
    \citet{liu-etal-2022-knowledge} & A/B & relevance matching \\ \hline
    \citet{rubin-etal-2023-entity} & A/B & virtual assistant \\ \hline
    \citet{dai-etal-2024-contrastive} & A/B & machine translation \\ \hline
    \citet{mohankumar-etal-2024-improving} & A/B & selecting bids for spons search \\ \hline
    \citet{chen-etal-2024-identifying} & observational & identifying important searches \\ \hline
    \citet{nygaard-etal-2024-news} & observational & credit risk monitoring \\ \hline
    \end{tabular}

    \caption{Impact papers found in ACL Anthology, where the description of the impact study in the paper was at least 0.5 pages and included at least one figure or table.}\label{tab-papers}

\end{table}

\section{Discussion}

The above literature survey suggests that papers that report real-world impact are extremely rare in the ACL Anthology, and even when they do appear, the impact study is often described very briefly and is treated a supplement to a metric study.  I personally find a reduction in real-world delivery defects to be far more impressive than precision/recall/accuracy metrics, but clearly \citet{maheshwary-etal-2024-pretraining} have the opposite opinion  (Section~\ref{sec:industry}).
In short, the ACL community seems to have little interest in evaluating real-world impact of NLP systems.

\revised{Note that impact evaluation is related to \textit{ethical evaluation}, since the underlying purpose of ethical evaluations is to understand whether NLP systems can harm people or society.  This is best measured by deploying
%Doctors, lawyers, teachers, and regulators are partially motivated by \textit{ethical} concerns.  They do not want NLP systems to harm people, and they realise that the only way to really assess whether systems are harmful is to deploy
systems and measuring harm-related KPIs, since harmful behaviour often occurs when NLP systems are used in complex real-world contexts which their developers did not anticipate.  From this perspective, impact evaluation has links to attempts to understand ethical considerations, such as \citet{mohammad-2022-ethics} and \citet{karamolegkou-etal-2025-ethical}.}

\subsection{When is impact evaluation appropriate?}

\revised{I do not expect all NLP papers to include an impact evaluation!  In medicine, impact evaluations are mainly used in papers which are intended to guide real-world decision making; they are not used in theoretical, modelling, and speculative papers, for example. 
Similarly in NLP,
%impact evaluation is not appropriate for many of the areas listed in the ARR Call for Papers, including Cultural Analytics, Cognitive Modelling, and Resources
I expect to see impact evaluations (at least some of the time) in papers that propose NLP models or systems for real-world usage; these papers are often (but not always) part of Industry or Applications tracks.  Impact evaluation is not appropriate for papers that are not applied and focus on theoretical issues, modelling fundamentals, cognitive science, etc.
I also do not expect to see impact evaluation in speculative, work-in-progress, or position papers.}

\revised{Perhaps most fundamentally,  impact evaluation should be part of the NLP research \scare{ecosystem} and happen at least in some cases; this will provide NLP as whole with feedback on the real-world utility of different approaches.}

%I do not expect to see an impact evaluation in papers that focus on theoretical, mathematical, or linguistic fundamentals;
%But papers that propose NLP models or systems for real-world usage should include impact evaluations in at least some cases; in other words, such papers should be part of the NLP research \scare{ecosystem}.
%Unfortunately, this is not the case in NLP in 2025.

%Impact evaluations are not appropriate for many types of papers, including theoretical papers, resource papers, position papers, and papers focusing on linguistics or cognitive science.  Impact evaluations are also not appropriate for preliminary and speculative work; indeed in medicine they are mainly used in papers which are intended to guide real-world decision making.  But I find it discouraging
%When looking at papers that show good performance at NLP tasks and use cases, its fine to have a mix where some papers use test set evaluation, and others using impact evaluation.  Perhaps we could have a similar pattern to medicine, where preliminary and speculative papers do not evaluate impact, but papers which are intended to guide real-world decision making are expected to include clinical trials or other types of impact evaluation.

\subsection{Barriers to impact evaluation}
\revised{There are barriers to impact evaluation in NLP.  One is simply that many NLP researchers have little knowledge of impact evaluation;  this is discussed in Section~\ref{encouraging}.  Practicalities can also be a barrier, including duration and cost of evaluation, and the in many cases the need to include a partner who can deploy a system and measure its impact.  These problems are less of an issue for companies, which may be why 71\% of the impact evaluations I found were in Industry Tracks.} 

\revised{Perhaps more fundamentally, the NLP's community lack of knowledge and interest in real-world impact evaluation may reflect its culture and the strong influence of machine learning (and perhaps the DARPA Common Task Method \cite{liberman2020human}), which focus on evaluations based on test sets.  Impact evaluations usually do not evaluate multiple systems on the same test set (Section~\ref{testset}), since this is very difficult to do when evaluating real-world usage of systems; hence they do not fit the test-set evaluation model.}

%Another barrier to impact evaluation is that knowledge and awareness of impact evaluation is poor in the NLP community, this is discussed in Section~\ref{encouraging}.

\subsection{Users want impact evaluation}

Communities which use NLP technology are frustrated by the lack of impact evaluation.  For example, \citet{hiebel2025clinical} review of NLG in clinical contexts states that \lingform{High scores on output quality metrics do not necessarily imply that deploying the system will benefit users.}  This point was also made very strongly to me by someone involved in assessing AI systems for use in part of the UK National Health Service; the lack of proper effectiveness evaluations (ideally randomised controlled clinical trials) is a major barrier in deploying AI systems in the NHS.

In the legal community, \citet{Kapoor_Henderson_Narayanan_2024} state that \lingform{the kinds of legal
applications we can legitimately use AI for should be determined
by the evaluations that reflect these uses of AI
in the real world.}  I am not aware of \emph{any} studies of real-world impact of NLP in legal contexts which have been published in the ACL Anthology.  Lawyers may be more willing than doctors to use NLP without strong evaluation, but they are concerned that they will lose their license if they use misbehaving AI\footnote{\url{https://www.cbsnews.com/colorado/news/colorado-lawyer-artificial-intelligence-suspension/}}.
Solid real-world evaluation will increase adoption and will also provide crucial feedback to NLP researchers working on AI in Law.

Concerns have also been expressed in other communities.  For example, in education, \citet{ganesh-etal-2023-mind} point out that the performance of an educational dialogue system is much worse in real classrooms.

Regulators are also aware of the need to test and evaluate AI systems in real-world usage.   Article 60 of the EU Artificial Intelligence Act\footnote{\url{https://artificialintelligenceact.eu/article/60/}} waives many of the Act's restrictions when AI systems are being tested in real-world usage.

In short, it is important to evaluate the real-world impact of NLP solutions, and this is acknowledged by both users and regulators.
Of course, the NLP community could leave this task to non-NLP researchers.  However, if we truly want to develop useful technology, it is much better if we are involved in the evaluations, and also if results from these evaluations are fed back to the NLP research community.

\section{Encouraging more impact evaluation}\label{encouraging}
\revised{I would love to see more impact evaluations in NLP papers!  Achieving this will not be easy, since it requires changing the research culture of NLP to focus more on real-world impact and less on SOTA-chasing, and changing research culture is difficult and also slow.  However, I do think we can make progress via better \textit{education} and \textit{incentives}.}

\revised{\textit{Education}: Many NLP researchers and PhD students have told me that they do not know how to evaluate impact; they were not taught this and have not previously thought about it.  I hope that this paper will help raise awareness of impact evaluation, including examples of how it is done; I also discuss impact evaluation in a recent book on NLG \cite{reiter2024}.  It would be very useful to have more education and awareness raising about impact evaluation, perhaps via tutorials, panel discussions, and/or keynotes at major NLP conferences.}

\revised{\textit{Incentives:} Researchers of course respond to incentives, and publication venues and/or funders could provide incentives to encourage impact evaluation.  For example, conferences and other venues could have special tracks or themes about impact, perhaps initially within Industry Tracks.  Likewise funders could have special programmes or calls which require impact evaluations, perhaps initially in domains such as medicine where this is especially important.}

\section{Conclusion}
If we want NLP technology to be used and genuinely help people, we need to understand its real-world impact on KPIs in deployed production usage.  Insights from this will help us develop more useful technology; in some fields (especially medicine), they are also an essential prerequisite to large-scale adoption.

Unfortunately the ACL community currently shows minimal interest in evaluating real-world impact.  I do not expect every paper in the Anthology to include an impact evaluation, but the number should be more than 0.1\%!
It is also frustrating that even when impact studies are done for commercial reasons, they are usually treated as being secondary to metric evaluations and also are usually not properly described.

Our community has developed amazing technology which has great potential to change the world; serious evaluation of its real-world impact will make NLP technology more useful and encourage more people to use it.

\section*{Acknowledgements}
\revised{My thanks to Liesl Osman and Scott Lennox for introducing me to impact evaluation and clinical trials, back in the 1990s.  Also thanks to my students and colleagues over the years who helped carry out impact-related evaluations, and to the many people who have discussed impact evaluation with me.  I would also like to thank Simone Balloccu, Daniel Braun, Nava Tintarev, and the members of the current Aberdeen NLP group for their very useful feedback about this paper, and the reviewers for their many helpful suggestions.}

\bibliographystyle{compling}
\bibliography{impactbib}

\end{document}